%% file: main.tex
\documentclass[10pt,twocolumn,letterpaper]{article}

\usepackage{cvpr}              
\input{preamble}
\definecolor{cvprblue}{rgb}{0.21,0.49,0.74}
\usepackage[pagebackref,breaklinks,colorlinks,allcolors=cvprblue]{hyperref}
\usepackage[most]{tcolorbox}
\tcbuselibrary{listings,breakable}

\newtcblisting{promptbox}[2][]{
breakable,
enhanced,
colback=gray!4,
colframe=black,
arc=2mm,
boxrule=0.8pt,
left=2mm,
right=2mm,
top=1mm,
bottom=1mm,
title=\textbf{#2},
fonttitle=\bfseries,
listing only,
listing options={
basicstyle=\ttfamily\footnotesize,
breaklines=true,
columns=fullflexible,
keepspaces=true,
showstringspaces=false
},
#1
}

\def\confName{CVPR}
\def\confYear{2026}

\title{ChartLens: A Dual-Branch Framework for Chart Data Correction and Factual Summary Refinement}

\author{
Hao Liu$^1$ \quad
Ruping Cao$^1$ \quad
Kun Wang$^1$ \quad
Zhiran Li$^1$ \quad
Fan Liu$^2$ \quad
Yupeng Hu$^1$ \quad
Liqiang Nie$^3$ \\
\\
$^1$Shandong University
$^2$Southeast University
$^3$Harbin Institute of Technology (Shenzhen) \\
{\tt\small \{liuh90210, caoruping657, khylon.kun.wang, zhiranli325,  liufancs, nieliqiang\}@gmail.com} \\
{\tt\small huyupeng@sdu.edu.cn}}

\begin{document}
\maketitle
\input{sec/0_abstract}    
\input{sec/1_intro}

\input{sec/2_method}

\input{sec/3_experiment}

{
    \small
    \bibliographystyle{ieeenat_fullname}
    \bibliography{main}
}

\input{sec/4_suppl}

\end{document}

%% file: sec/0_abstract.tex
\begin{abstract}
In this report, we present our champion solution for the DataMFM Challenge Track 2: Chart Understanding. This track requires models to recover structured chart data and generate faithful natural-language summaries from chart images. To address the complementary requirements of accurate data extraction and factual narration, we propose \textbf{ChartLens}, a dual-branch framework for chart data correction and summary refinement. ChartLens consists of two key modules: Structure-Aware CSV Verification and Correction (SAVC) and Text-Retention-Guided Summary Refinement (TRSR). SAVC improves the reliability of structured data extraction through verification and correction, while TRSR enhances summary generation by preserving critical textual and numerical evidence from charts. By combining model adaptation, correction-based generation, and OCR-assisted evidence grounding, ChartLens improves both structured data recovery and summary factuality. On the test set, our final system achieves an overall score of 69.10 and ranks \textbf{first} in Track 2, demonstrating its effectiveness for accurate chart understanding. Our code will be released at: \url{https://github.com/iLearn-Lab/CVPRW26-ChartLens}.
\end{abstract}

%% file: sec/1_intro.tex
\section{Introduction}

With the increasing prevalence of data-driven documents, charts have become a key medium for communicating numerical information across diverse domains~\cite{kondic2026chartnet, huang2024pixels}. However, the diverse visual layouts and compact semantic structures of charts make it challenging to automatically recover the underlying data and describe it faithfully~\cite{ouyang2025omnidocbench, wang2024charxiv, li2026unim}. Unlike general multimodal understanding that primarily focuses on semantic recognition or image-text alignment~\cite{wang2024explicit, liu2025gaming, hu2026glance, wang2025redundancy, li2025mist, hu2026visual, xiang2025dkdm, xiang2026tina}, chart understanding requires models to jointly interpret visual encodings, recover structured numerical relationships, and generate fact-grounded textual descriptions. Consequently, Chart Understanding has emerged as a crucial research problem in multimodal document intelligence. Specifically, chart understanding aims to transform chart images into machine-readable structured data and evidence-grounded natural-language summaries~\cite{xu2025chartmoe, zhao2025chartcoder}. This task facilitates numerous applications, such as automated document parsing, data analysis, and evidence-grounded information retrieval~\cite{dong2026doc, li2026efficient, wang2026cross, li2025dcount, liu2025curmim, hu2021video, hu2021coarse, zhao2023cogcn, hu2023semantic}.

\begin{figure}[t]
\centering
\includegraphics[width=0.46\textwidth]{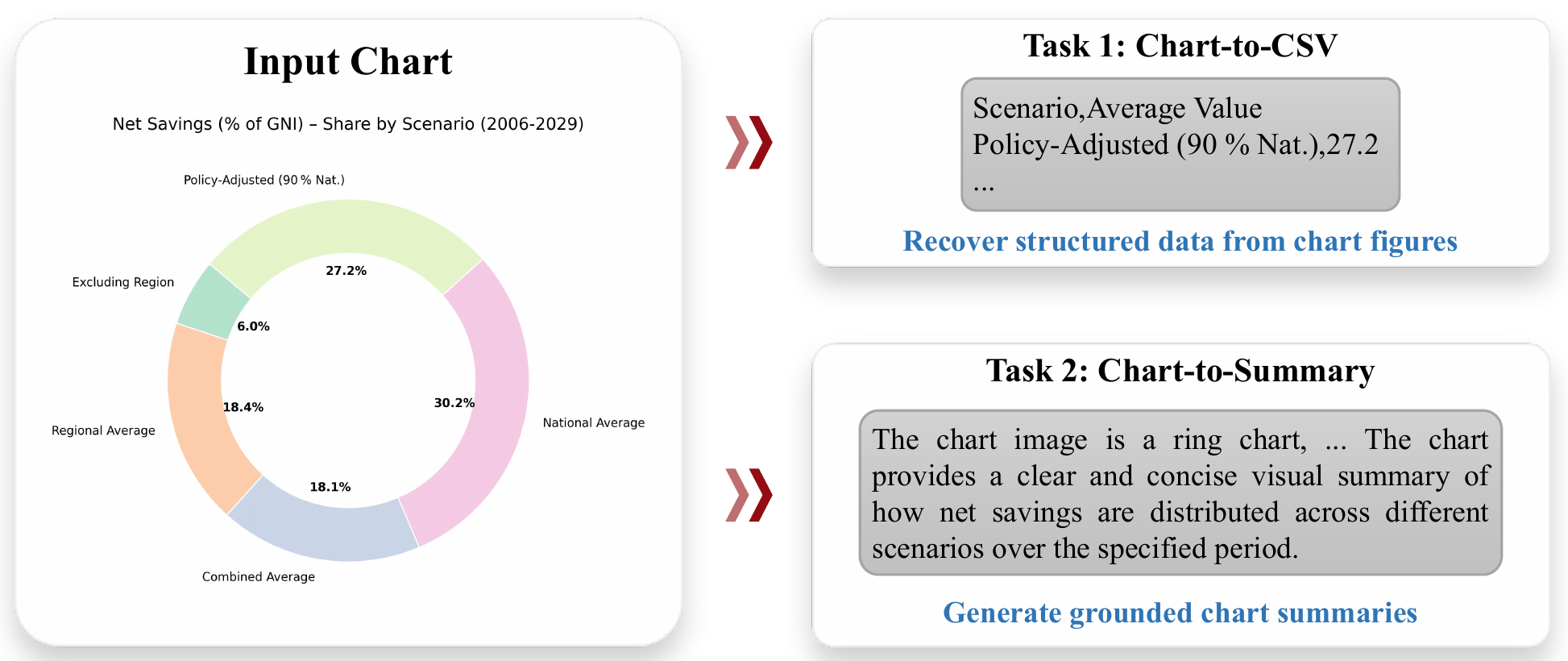}
\vspace{-2mm}
\caption{Task definition of DataMFM Challenge Track 2. Given an input chart image, the model is required to recover structured data through chart-to-CSV extraction and generate grounded chart summaries through chart-to-summary generation.}
\vspace{-6mm}
\label{fig:task_definition}
\end{figure}

The DataMFM Challenge Track 2\footnote{https://datamfm.github.io/challenge.html} focuses on chart understanding in multimodal document scenarios. As shown in Figure~\ref{fig:task_definition}, given an input chart image, participating models are required to complete two complementary tasks. The first task is \textit{\textbf{chart-to-CSV extraction}}, which aims to recover structured data from chart figures. The second task is \textit{\textbf{chart-to-summary generation}}, which requires the model to generate grounded chart summaries. Therefore, this track is not only about recognizing chart content, but also about organizing visual and numerical evidence into reliable structured and textual outputs.

\begin{figure}[t]
\centering
\includegraphics[width=0.46\textwidth]{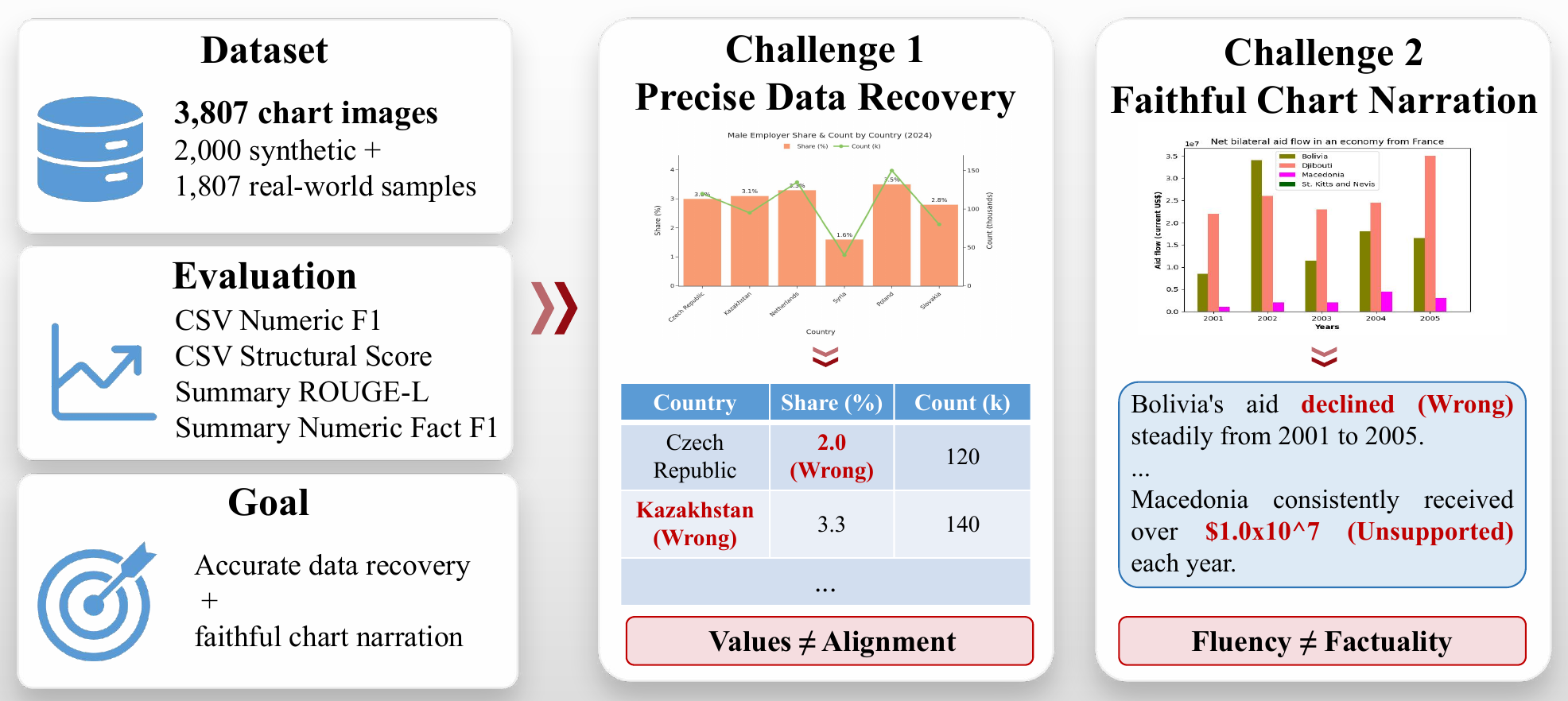}
\vspace{-2mm}
\caption{DataMFM Track 2 evaluates chart understanding with 3,807 chart images and multiple metrics for CSV extraction and summary generation. The key challenges are precise data recovery, where values must be correctly aligned with chart structures, and faithful chart narration, where fluent summaries must remain factually grounded.}
\vspace{-6mm}
\label{fig:motivation}
\end{figure}

Despite the strong visual-language capabilities of recent multimodal foundation models~\cite{radford2021learning, li2022blip, li2023blip}, directly generating chart outputs remains unreliable for this challenge. The core limitation is that chart understanding requires both accurate data recovery and faithful chart narration, while direct generation can easily violate either of them. As illustrated in Figure~\ref{fig:motivation}, the benchmark contains 3,807 chart images, including both synthetic and real-world samples, and evaluates models from both structured extraction and summary generation perspectives. This evaluation setting exposes two practical challenges: 
1) \textbf{Precise Data Recovery.} Chart-to-CSV extraction requires more than recognizing numerical values. A value is meaningful only when it is correctly aligned with its corresponding category, legend, axis, or column. In other words, values are not equivalent to alignment. A model may read a number correctly but place it under the wrong group or field, resulting in an incorrect CSV even when the local value appears plausible. 
2) \textbf{Faithful Chart Narration.} Chart-to-summary generation requires more than producing fluent language. A summary can be coherent and readable while still introducing unsupported trends or hallucinated numerical claims. In other words, fluency is not equivalent to factuality. These two challenges indicate that chart understanding should not be treated as a one-shot generation problem.

To tackle these challenges, we adopt a verification-guided correction strategy that jointly improves structured data recovery and factual chart narration. The core intuition is that raw multimodal outputs should not be directly used as final predictions; instead, they should be verified and corrected according to chart structures and textual evidence. To be specific, we propose \textbf{ChartLens}, a dual-branch framework for chart data correction and factual summary refinement. Unlike direct generation pipelines that rely on a single model to produce final outputs, ChartLens decomposes chart understanding into two complementary branches. 1) \textbf{Structure-Aware CSV Verification and Correction} (SAVC) verifies the generated CSV from the perspectives of structure, completeness, and numerical consistency, and corrects unreliable headers, categories, legends, or values when necessary. 2) \textbf{Text-Retention-Guided Summary Refinement} (TRSR) uses chart textual cues to guide summary refinement, encouraging the generated summary to retain key titles, legends, annotations, and numerical evidence. Together, SAVC and TRSR form a unified correction-oriented framework that preserves reliable predictions while revising structurally inconsistent or factually unsupported content.

In summary, our main contributions are threefold:
\begin{itemize}[leftmargin=*]
\item We propose \textbf{ChartLens}, a dual-branch framework for chart understanding that improves both structured data recovery and faithful summary generation through verification-guided correction.

\item We design two complementary branches for the two subtasks: SAVC for correcting chart-derived tables, and TRSR for refining summaries with chart textual and numerical evidence.

\item We demonstrate the effectiveness of the proposed framework on the DataMFM Challenge Track 2, where our final model achieves an overall score of 69.10 and ranks \textbf{first} among submitted solutions.
\end{itemize}

%% file: sec/2_method.tex
\section{Method}
\begin{figure*}[t]
\centering
\includegraphics[width=0.7\textwidth]{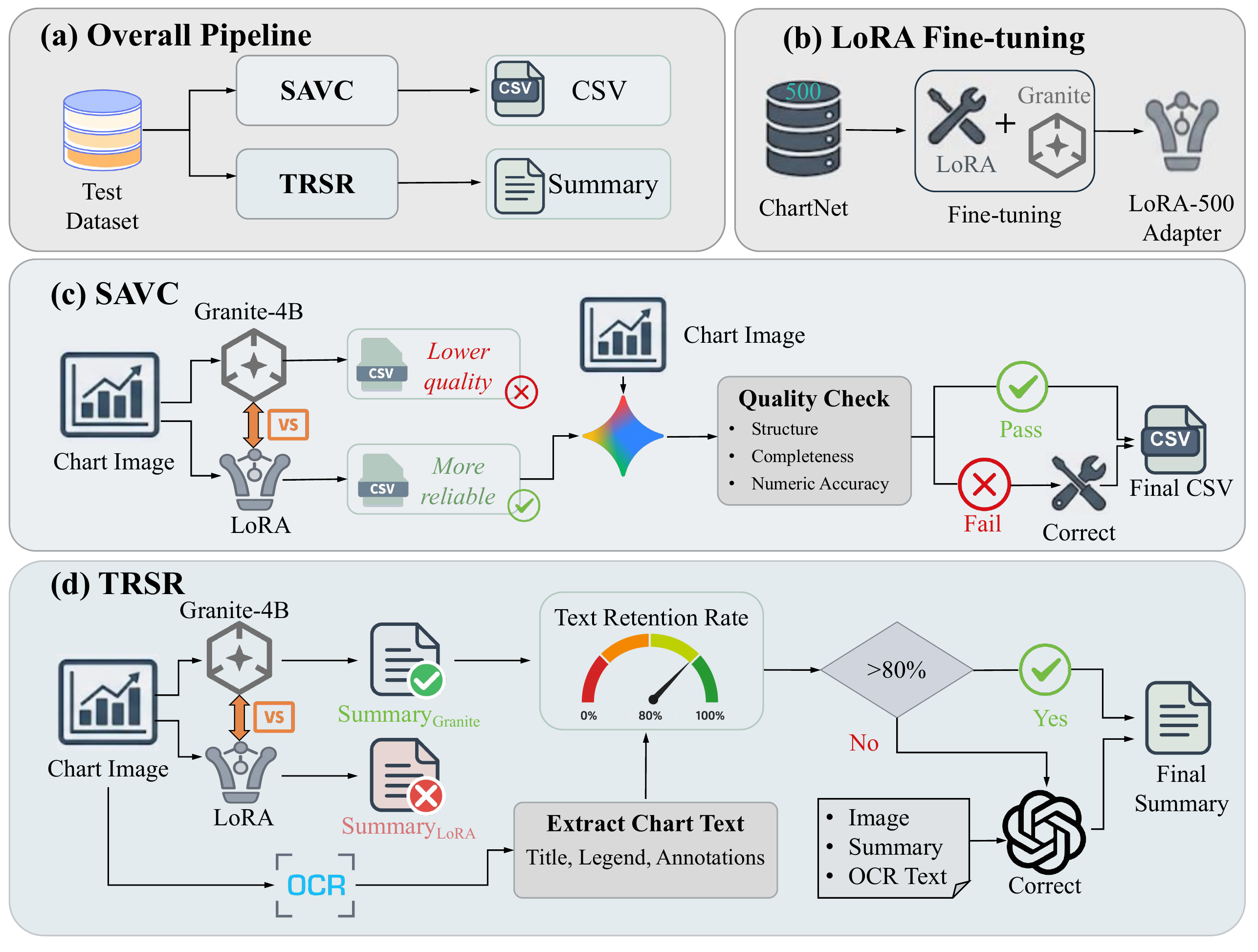}
\vspace{-2mm}
\caption{Overview of ChartLens. (a) The pipeline uses parallel branches for CSV generation and summary generation. (b) Granite-Vision-4.1-4B is adapted with LoRA for CSV initialization. (c) SAVC verifies and corrects the generated CSV through structure, completeness, and numerical accuracy checks. (d) TRSR evaluates text retention using OCR-extracted chart text and refines summaries when necessary.}
\vspace{-4mm}
\label{fig:method_overview}
\end{figure*}

In this section, we first formulate the chart understanding task and present the overall architecture of ChartLens, as illustrated in Figure~\ref{fig:method_overview}. We then describe how initial CSV and summary outputs are constructed from multimodal foundation models. After that, we introduce the two correction branches, namely SAVC and TRSR. Finally, we summarize the overall inference pipeline used for the final submission.

\subsection{Task Formulation}

Given a chart image $I$, DataMFM Track 2 requires the model to produce two outputs: a structured CSV table $C$ and a natural-language summary $S$. The prediction target is therefore the pair $(C,S)$.

For chart-to-CSV extraction, $C$ should recover the underlying chart data:
\begin{equation}
C=\{H,R,V\},
\end{equation}
where $H$ denotes headers or fields, $R$ denotes category entries, and $V$ denotes numerical values. A correct CSV should not only preserve visible values, but also place them within the correct chart structure.

For chart-to-summary generation, the output $S$ should provide a concise textual description of the chart. Different from open-ended captioning, $S$ must be grounded in the visual and numerical evidence of $I$, avoiding unsupported trends or hallucinated values. Therefore, the task requires the model to jointly optimize structural correctness in $C$ and factual consistency in $S$.

\subsection{Initial Output Construction}

We construct initial predictions with a fixed multimodal backbone rather than treating model selection as part of the method. Specifically, we use Granite-Vision-4.1-4B~\cite{granite2026} as the base generator to obtain the initial chart outputs:
\begin{equation}
(C_{\text{base}},S_{\text{base}})=M_{\text{base}}(I),
\end{equation}
where $C_{\text{base}}$ and $S_{\text{base}}$ denote the directly generated CSV and summary, respectively. The corresponding prompt is provided in the supplementary material.

As shown in Figure~\ref{fig:method_overview}(b), we first adapt Granite-Vision-4.1-4B with LoRA~\cite{hulora} on the released ChartNet~\cite{kondic2026chartnet} training data to strengthen its structured extraction ability:
\begin{equation}
C_{\text{LoRA}}=M_{\text{LoRA}}(I).
\end{equation}
Since this adaptation mainly improves structured data recovery, we use the LoRA-generated CSV as the initial table and retain the base summary as the initial summary:
\begin{equation}
(C_0,S_0)=(C_{\text{LoRA}},S_{\text{base}}).
\end{equation}
The resulting pair $(C_0,S_0)$ serves as the starting point for the subsequent SAVC and TRSR branches.

\subsection{SAVC}

As shown in Figure~\ref{fig:method_overview}(c), SAVC aims to correct the initial CSV $C_0$ by verifying its consistency with the input chart. We formulate this branch as:
\begin{equation}
C^\ast=\Phi_{csv}(I,C_0,P_{csv}),
\end{equation}
where $\Phi_{csv}$ denotes the correction model, and $P_{csv}$ denotes the verification prompt provided in the supplementary material.

Following the task formulation, a reliable CSV should preserve both chart values and their structural correspondence. Therefore, SAVC verifies the initial CSV from three perspectives: 1) \textbf{structural consistency}, which checks whether the table follows the chart organization; 2) \textbf{content completeness}, which checks whether visible entries are sufficiently preserved; and 3) \textbf{numerical accuracy}, which checks whether extracted values are consistent with the chart evidence. These checks guide the correction model to focus on structural shifts, missing fields, wrong value assignments, and unsupported entries.

Instead of regenerating the full table from scratch, SAVC performs edit-based correction:
\begin{equation}
C^\ast=C_0\oplus\Delta C,
\end{equation}
where $\Delta C$ denotes the predicted correction operation and $\oplus$ denotes applying the edit to the initial CSV. This design preserves correct entries in $C_0$ while revising unreliable parts. In our implementation, Gemini-3.5-Flash~\cite{gemini35flashmodelcard} is used as the correction model to verify the chart image and the initial CSV jointly, thereby reducing pattern-driven structural hallucination in chart-to-CSV extraction.

\subsection{TRSR}

As shown in Figure~\ref{fig:method_overview}(d), TRSR improves the factual grounding of the initial summary $S_0$ by introducing OCR-derived text. Given the image $I$, we extract textual cues:
\begin{equation}
T=\mathrm{OCR}(I)=\{t_1,t_2,\ldots,t_N\},
\end{equation}
where $t_i$ denotes a detected text span from the chart.

To determine whether refinement is necessary, TRSR estimates how much key chart text is retained by the initial summary. Let $\mathcal{K}(\cdot)$ denote the extraction of key textual units. The text-retention score is defined as:
\begin{equation}
\rho(S_0,T)=
\frac{|\mathcal{K}(S_0)\cap\mathcal{K}(T)|}
{|\mathcal{K}(T)|}.
\end{equation}
If the retention score is sufficient, the initial summary is preserved. Otherwise, TRSR refines it with the chart image, the initial summary, and OCR evidence:
\begin{equation}
S^\ast=
\begin{cases}
S_0, & \rho(S_0,T)\geq\tau,\\
\Phi_{sum}(I,S_0,T,P_{sum}), & \rho(S_0,T)<\tau,
\end{cases}
\end{equation}
where $\Phi_{sum}$ denotes the summary refinement model implemented with GPT-5.5~\cite{openai2026gpt55systemcard}, $\tau$ is the retention threshold, and the summary refinement prompt $P_{sum}$ is provided in the supplementary material.

OCR is used as an auxiliary factual anchor rather than a replacement for visual reasoning. When the initial summary misses important chart text, OCR provides explicit evidence for recovery. When the summary contains unsupported trends or incomplete numerical descriptions, TRSR encourages the model to revise the statement according to visible chart evidence. This improves factual consistency while maintaining the fluency of the original summary.

\subsection{Overall Inference}

During inference, ChartLens follows the staged process illustrated in Figure~\ref{fig:method_overview}. For each chart image $I$, we first construct the initial outputs $(C_0,S_0)$ using the LoRA-adapted CSV generator and the selected direct summary generator. Then, SAVC takes $C_0$ as input and produces the final corrected CSV $C^\ast$ through structure-aware verification and correction. In parallel, TRSR takes $S_0$ and OCR-extracted chart text $T$ as input, and produces the final refined summary $S^\ast$ according to text-retention guidance. The resulting pair $(C^\ast,S^\ast)$ is used as the prediction for the chart and converted into the official JSONL format for submission.

%% file: sec/3_experiment.tex
\section{Experiments}

\subsection{Experimental Settings}

\textbf{Dataset.}
DataMFM Challenge Track 2 is built on a newly prepared chart understanding dataset based on ChartNet~\cite{kondic2026chartnet}. The dataset contains 3,807 chart images, including 2,000 synthetic samples and 1,807 real-world samples. This track focuses on two chart understanding tasks: chart-to-CSV and chart-to-summary. The former requires the model to convert a chart image into a structured CSV table, while the latter requires the model to generate a textual summary that accurately reflects the visual and numerical information in the chart.

\textbf{Evaluation Metrics.}
Following the official evaluation protocol of DataMFM Challenge Track 2, we evaluate each submission using four metrics: CSV Numeric F1, CSV Structural Score, Summary ROUGE-L, and Summary Numeric Fact F1. The final Overall score is computed by the official evaluation script.

\textbf{Implementation Details.}
We use Granite-Vision-4.1-4B~\cite{granite2026} as the base multimodal generator and adapt it with LoRA~\cite{hulora}. The LoRA model is trained on the released ChartNet-based training data. The LoRA rank is set to 16, the LoRA scaling factor is set to 32, the learning rate is set to 1e-4, the batch size is set to 2, and the number of training epochs is set to 2.0. For TRSR, we use PaddleOCR~\cite{cui2025paddleocr} to extract chart text and set the text-retention threshold to $\tau=0.8$. All experiments are conducted on a single NVIDIA RTX 4090 GPU. The prompts used for direct generation, CSV correction, and summary refinement are provided in the supplementary material.

\subsection{Leaderboard Comparison}

Table~\ref{tab:leaderboard} reports the official leaderboard comparison of DataMFM Challenge Track 2. Our team, iLearn-Chart, ranks first among all submitted solutions with an Overall score of 69.10. Compared with the second-ranked team, our solution improves the Overall score by 1.53 points. In terms of sub-metrics, our model achieves the best CSV Numeric F1, CSV Structural Score, and Summary Numeric Fact F1, indicating its advantage in both structured value recovery and factual summary generation.

\begin{table}[t]
\centering
\caption{Leaderboard on DataMFM Challenge Track 2. $N$, $S$, $L$, and $F$ denote CSV Numeric F1, CSV Structural Score, Summary ROUGE-L, and Summary Numeric Fact F1, respectively.}
\vspace{-2mm}
\label{tab:leaderboard}
\resizebox{0.99\linewidth}{!}{%
\begin{tabular}{lccccc}
\toprule
Team & $N$ & $S$ & $L$ & $F$ & Overall \\
\midrule
hskl18 & 47.29 & 37.77 & 20.17 & 51.71 & 39.23 \\
SFD & 65.37 & 68.55 & 14.51 & 36.26 & 46.17 \\
HHHHHHHHHH & 67.66 & 66.81 & 35.65 & 70.76 & 60.22 \\
Wind\_Rain\_Tower & 76.22 & 73.52 & 29.47 & 62.27 & 60.37 \\
ytttttt & 73.43 & 71.57 & \textbf{45.69} & 73.48 & 66.04 \\
anmspro & 75.26 & 74.65 & 44.81 & 73.73 & 67.11 \\
acceed & 76.28 & 73.63 & 45.23 & 73.69 & 67.21 \\
durgasandeep & 76.03 & 74.78 & 45.30 & 73.95 & 67.52 \\
Zhiheng & 76.55 & 74.82 & 45.23 & 73.69 & 67.57 \\ 
\midrule
\textbf{iLearn-Chart (Ours)} & \textbf{80.62} & \textbf{75.66} & 45.57 & \textbf{74.55} & \textbf{69.10} \\
\bottomrule
\end{tabular}%
}
\vspace{-4mm}
\end{table}

\begin{table}[t]
\centering
\caption{Candidate model comparison. $N$, $S$, $L$, and $F$ denote CSV Numeric F1, CSV Structural Score, Summary ROUGE-L, and Summary Numeric Fact F1, respectively. ``FT" denotes the fine-tuned model.}
\label{tab:candidate_model}
\vspace{-2mm}
\resizebox{0.99\linewidth}{!}{%
\begin{tabular}{lccccc}
\toprule
Model & $N$ & $S$ & $L$ & $F$ & Overall \\
\midrule
Gemini-3.5-Flash & \textbf{79.45} & 74.07 & \textbf{44.96} & 71.17 & 67.41 \\
MiMo-v2.5 & 73.68 & 69.98 & 35.06 & 65.99 & 61.18 \\ 
Granite-Vision-4.1-4B & 75.69 & 74.72 & \textbf{44.96} & \textbf{73.19} & 67.14 \\
Granite-Vision-4.1-4B (FT) & 79.13 & \textbf{75.94} & 44.25 & 72.29 & \textbf{67.90} \\
\bottomrule
\end{tabular}}

\end{table}

\subsection{Candidate Model Comparison}

We first compare several multimodal foundation models under direct generation. This experiment is used to analyze the capability of different backbones, rather than treating model selection as part of the proposed method. As shown in Table~\ref{tab:candidate_model}, Gemini-3.5-Flash achieves strong CSV Numeric F1, while Granite-Vision-4.1-4B provides competitive summary factuality. After LoRA adaptation, Granite-Vision-4.1-4B improves CSV Numeric F1 from 75.69 to 79.13 and CSV Structural Score from 74.72 to 75.94, showing that LoRA adaptation is beneficial for structured data recovery.

\subsection{Fine-Tuning Data Size}

We then study whether increasing the LoRA fine-tuning scale consistently improves performance. As shown in Table~\ref{tab:scale}, fine-tuning with 500 images achieves an Overall score of 67.90, while increasing the fine-tuning data size to 10K images slightly decreases the Overall score to 67.67. Although the larger setting improves Summary Numeric Fact F1 from 72.29 to 72.71, it reduces CSV Numeric F1 and Summary ROUGE-L. This suggests that simply increasing the fine-tuning scale does not necessarily improve the final performance. A possible reason is that the adapted model may overfit specific chart patterns or generation styles, weakening its generalization to the evaluation distribution.

\begin{table}[t]
\centering
\caption{Fine-tuning validation for Granite-Vision-4.1-4B. $N$, $S$, $L$, and $F$ denote CSV Numeric F1, CSV Structural Score, Summary ROUGE-L, and Summary Numeric Fact F1, respectively.}
\vspace{-2mm}
\label{tab:scale}
\resizebox{\linewidth}{!}{%
\begin{tabular}{lccccc}
\toprule
Model & $N$ & $S$ & $L$ & $F$ & Overall \\
\midrule
Fine-tuned Granite-Vision-4.1-4B (500 images) & \textbf{79.13} & \textbf{75.94} & \textbf{44.25} & 72.29 & \textbf{67.90} \\
Fine-tuned Granite-Vision-4.1-4B (10K images) & 78.62 & 75.92 & 43.43 & \textbf{72.71} & 67.67 \\
\bottomrule
\end{tabular}}
\vspace{-4mm}
\end{table}

\subsection{Generation Strategy and OCR Ablation}

Table~\ref{tab:Generation} reports the effect of correction-based generation and OCR-assisted summary refinement. The selected direct-output setting achieves an Overall score of 68.30. After applying Gemini-3.5-Flash as the verification and correction model, the Overall score increases to 68.40. The improvement mainly comes from CSV Numeric F1, which increases from 79.13 to 80.62, and Summary ROUGE-L, which increases from 44.96 to 45.09. This demonstrates that correcting generated outputs is more effective than directly trusting raw predictions.

We further introduce OCR cues for summary refinement. Compared with correction without OCR, adding OCR improves Summary ROUGE-L from 45.09 to 45.57 and Summary Numeric Fact F1 from 72.23 to 74.55. The CSV metrics remain unchanged because OCR is only used in the summary branch. With correction and OCR jointly applied, the final model achieves the best Overall score of 69.10.

\begin{table}[t]
\centering
\caption{Generation strategy and OCR ablation. $N$, $S$, $L$, and $F$ denote CSV Numeric F1, CSV Structural Score, Summary ROUGE-L, and Summary Numeric Fact F1, respectively.}
\vspace{-2mm}
\label{tab:Generation}
\resizebox{\linewidth}{!}{%
\begin{tabular}{lccccc}
\toprule
Setting & $N$ & $S$ & $L$ & $F$ & Overall \\
\midrule
Direct output & 79.13 & \textbf{75.94} & 44.96 & 73.19 & 68.30 \\
Correction without OCR & \textbf{80.62} & 75.66 & 45.09 & 72.23 & 68.40 \\
Correction with OCR & \textbf{80.62} & 75.66 & \textbf{45.57} & \textbf{74.55} & \textbf{69.10} \\
\bottomrule
\end{tabular}}
\end{table}

\section{Qualitative Analysis}

\subsection{Successful Case}

\begin{figure}[t]
\centering
\includegraphics[width=0.92\linewidth]{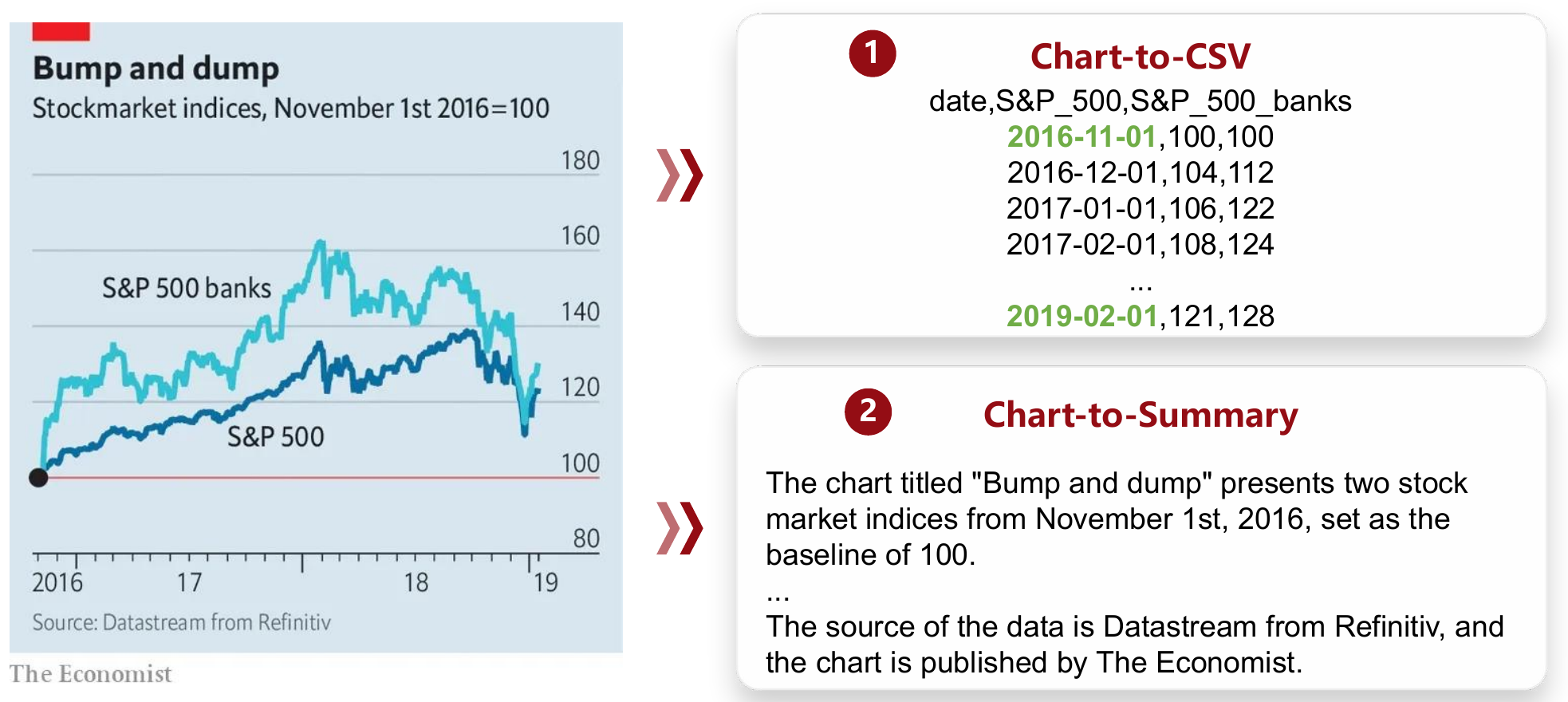}
\vspace{-2mm}
\caption{Successful case visualization. The correction strategy removes pattern-driven temporal hallucination and preserves the correct date range and series structure.}
\vspace{-4mm}
\label{fig:success}
\end{figure}

Figure~\ref{fig:success} presents a successful case from a time-series chart titled ``Bump and dump''. A common failure of direct generation is pattern-driven temporal hallucination. For example, because monthly time series often follow regular full-year patterns, the model may extend the data range to December 2019, even when the chart only ends in February 2019. In this case, SAVC uses Gemini-3.5-Flash as a quality checker rather than a pure generator. It verifies whether the initial CSV is consistent with the visible date range and series structure, and then corrects the original output accordingly. The corrected CSV stops at February 2019 and preserves the two series, namely S\&P 500 and S\&P 500 banks. The summary also retains key chart information, including the title, baseline setting, and data source. This case shows that correction-based generation can reduce temporal hallucination and improve factual consistency.

\begin{figure}[t]
\centering
\includegraphics[width=0.92\linewidth]{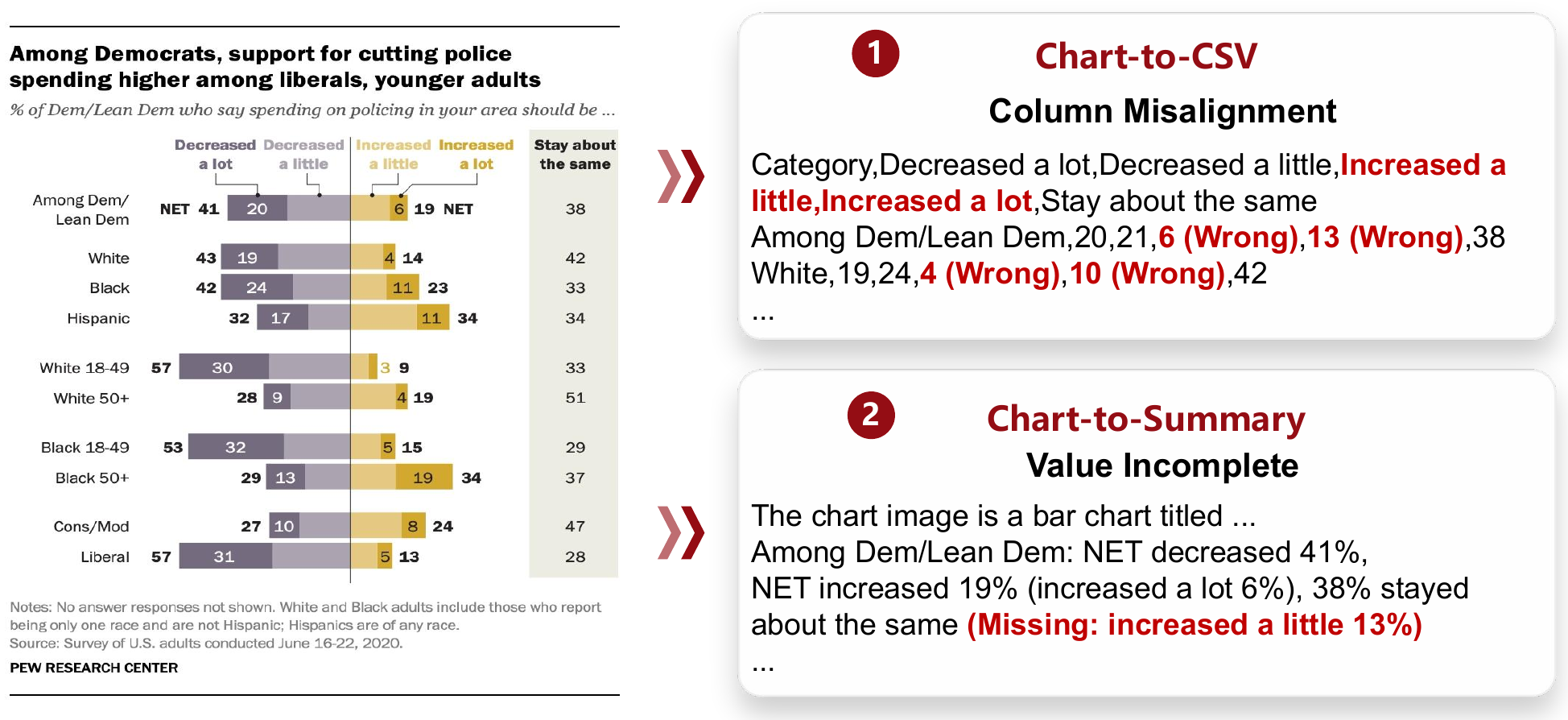}
\vspace{-2mm}
\caption{Failure case visualization. The model reads several numerical values but swaps two adjacent semantic columns, revealing the remaining challenge of structural alignment.}
\vspace{-4mm}
\label{fig:failure}
\end{figure}

\subsection{Failure Case}

Figure~\ref{fig:failure} shows a remaining failure case. The model does not fail at number recognition. Instead, it makes an alignment error between two adjacent semantic columns, namely ``Increased a little" and ``Increased a lot". For example, the value 6 should belong to ``Increased a lot", but the model assigns it to ``Increased a little". This error may come from pattern-driven structural reasoning: since the left side follows the order ``Decreased a lot" and then ``Decreased a little", the model incorrectly applies a similar order to the right side, although the actual visual layout is different. The generated summary partially inherits this alignment error and misses some sub-category values. This case indicates that chart understanding still requires stronger structural alignment, especially for charts with dense visual layouts or asymmetric category arrangements.

\section{External Resource Disclosure}

Our model uses the released ChartNet~\cite{kondic2026chartnet} training data for LoRA~\cite{hulora} adaptation and evaluation. We use pretrained multimodal foundation models, including Granite-Vision-4.1-4B~\cite{granite2026}, Gemini-3.5-Flash~\cite{gemini35flashmodelcard}, and GPT-5.5~\cite{openai2026gpt55systemcard}. Gemini-3.5-Flash is used as the verification and correction model in SAVC, while GPT-5.5 is used as the summary refinement model in TRSR. We also use PaddleOCR~\cite{cui2025paddleocr} to extract chart text for summary refinement. No additional manually annotated chart labels are introduced beyond the released challenge resources.

\section{Conclusion}

In this report, we present ChartLens, our champion solution for DataMFM Challenge Track 2. The core insight of ChartLens is that chart understanding should not be treated as pure direct generation. Instead, accurate chart understanding requires verification-guided correction over both structured data and textual narration. To this end, ChartLens combines LoRA-adapted CSV initialization, SAVC for structure-aware CSV correction, and TRSR for OCR-assisted summary refinement. Experimental results show that correction-based generation improves direct outputs, while OCR cues further enhance summary factuality. Our final model achieves an Overall score of 69.10 and ranks first in Track 2. Remaining failures, such as column misalignment, suggest that stronger schema reasoning and visual-structural alignment remain important directions for robust chart understanding.

%% file: sec/4_suppl.tex
\clearpage
\setcounter{page}{1}
\maketitlesupplementary


\setcounter{section}{0}

\section{Prompt Details}
\label{app:prompts}

This supplementary material provides the prompt templates used in ChartLens. Variables enclosed by braces, such as \texttt{{imagename}}, \texttt{{baseline\_csv}}, and \texttt{{ocr\_reference}}, are replaced with instance-specific inputs during inference.

\subsection{Direct Generation Prompt}

\begin{promptbox}{Direct Generation Prompt}
You are a chart understanding assistant.

You will receive one chart image. Your task is to generate:

1. A structured CSV table that recovers the chart data.
2. A concise natural-language summary grounded in the chart.

CSV requirements:

* The CSV must have a header row.
* Preserve visible categories, legends, series, and numerical values.
* Use plain numeric cells whenever possible.
* Put units in column names when needed.
* Do not invent invisible rows, columns, labels, sources, or values.

Summary requirements:

* The summary must be one paragraph.
* Describe the chart type, title, main variables, and key trends.
* Mention important numerical facts only when they are supported by the chart.
* Do not hallucinate unsupported trends, comparisons, or values.

Return only one valid JSON object:
{
"csv": "...",
"summary": "..."
}
\end{promptbox}

\subsection{SAVC Prompt}

\begin{promptbox}{SAVC Prompt}
You are a strict chart-to-CSV judge and repair expert for a Chart Understanding benchmark.

Image split: {split_name}
Image filename: {imagename}

You are given the chart image and candidate CSV outputs.
Your job:

1. Compare every candidate CSV against the image.
2. Select the candidate that best matches the image.
3. Improve the selected CSV when it has small or moderate mistakes.
4. If all candidates are clearly inconsistent with the image, regenerate a new CSV from the image.

Important decision rules:

* Do NOT simply copy the best candidate.
* If a candidate has the right table structure but a few values or labels are wrong, keep its structure and correct those values or labels.
* If a candidate misses visible categories or visible series, add them.
* If a candidate invents invisible categories or invisible series, remove them.
* Use printed values exactly when visible.
* If values are not printed, estimate carefully from the axis scale.
* Preserve original labels as closely as possible.

CSV formatting rules:

* Output clean CSV text only inside the JSON field "csv".
* The CSV must have a header row.
* Use one column per visible series when that is the natural chart structure.
* Use plain numeric cells whenever possible.
* Put units in column names when needed.
* If a field contains a comma, wrap the field in double quotes.
* Do not add explanations, notes, markdown, or code fences inside the CSV.
* Keep newline characters as \n in the JSON string.

Return only one valid JSON object:
{
"csv": "...",
"selected_candidate": "model name, or regenerate",
"action": "copy|repair|regenerate",
"reason": "brief reason in English"
}

Candidate CSVs:
{candidates_text}
\end{promptbox}

\subsection{TRSR Prompt}

\begin{promptbox}{TRSR Prompt}
You are repairing a chart summary for a Chart Understanding benchmark.

Important context:

* You are given the chart image.
* You are given the original predicted summary.
* You are also given OCR text extracted from the image.
* The OCR may contain errors and is only a reference.
* Your goal is to improve textual faithfulness while changing the original summary as little as possible.

Image filename:
{imagename}

Split:
{split_name}

OCR reference:
{ocr_reference}

Original summary:
{original_summary}

Repair rules:

1. Make the smallest possible edits to the original summary.
2. Preserve the original sentence order, paragraph structure, chart type, main trend descriptions, and numeric content unless the image clearly shows they are wrong.
3. If the chart contains visible text such as title, subtitle, axis label, legend label, category label, source, note, or publisher, use the exact wording from the chart.
4. Do NOT replace visible chart text with synonyms or expanded paraphrases.
5. Trust the image over OCR when they conflict.
6. Do not add invisible facts, invisible labels, or unsupported numeric values.
7. Do not rewrite the whole summary unnecessarily.

Return only one valid JSON object:
{
"summary": "..."
}

Final output requirements:

* The summary must be a single paragraph string.
* No Markdown.
* No code fence.
* No explanation outside JSON.
  \end{promptbox}